\documentclass[conference]{IEEEtran}
\IEEEoverridecommandlockouts
\usepackage{cite}
\usepackage{amsmath,amssymb,amsfonts}
\usepackage{graphicx}
\usepackage{textcomp}
\usepackage{xcolor}
\usepackage{xspace}
\usepackage[bookmarks=true]{hyperref}
\usepackage{listings}

\makeatletter
\let\old@lstKV@SwitchCases\lstKV@SwitchCases
\def\lstKV@SwitchCases#1#2#3{}
\makeatother
\usepackage{lstlinebgrd}
\makeatletter
\let\lstKV@SwitchCases\old@lstKV@SwitchCases

\lst@Key{numbers}{none}{%
\def\lst@PlaceNumber{\lst@linebgrd}%
\lstKV@SwitchCases{#1}%
{none:\\%
left:\def\lst@PlaceNumber{\llap{\normalfont
\lst@numberstyle{\thelstnumber}\kern\lst@numbersep}\lst@linebgrd}\\%
right:\def\lst@PlaceNumber{\rlap{\normalfont
\kern\linewidth \kern\lst@numbersep
\lst@numberstyle{\thelstnumber}}\lst@linebgrd}%
}{\PackageError{Listings}{Numbers #1 unknown}\@ehc}}
\makeatother

\usepackage{algorithm}
\usepackage{algorithmic}

\usepackage[nolist]{acronym}

\definecolor{dkgreen}{rgb}{0,0.6,0}
\definecolor{gray}{rgb}{0.5,0.5,0.5}
\definecolor{mauve}{rgb}{0.58,0,0.82}

\begin{document}

\title{Auptimizer - an Extensible, Open-Source Framework for Hyperparameter Tuning}

\author{\IEEEauthorblockN{Jiayi Liu} \IEEEauthorblockA{\textit{Advanced AI} \\
\textit{LG Electronics}\\
Santa Clara, CA, USA\\
Jason.Liu@lge.com}
\and
\IEEEauthorblockN{Samarth Tripathi} \IEEEauthorblockA{\textit{Advanced AI} \\
\textit{LG Electronics}\\
Santa Clara, CA, USA\\
Samarth.Tripathi@lge.com}
\and
\IEEEauthorblockN{Unmesh Kurup} \IEEEauthorblockA{\textit{Advanced AI} \\
\textit{LG Electronics}\\
Santa Clara, CA, USA\\
Unmesh.Kurup@lge.com}
\and
\IEEEauthorblockN{Mohak Shah} \IEEEauthorblockA{\textit{Advanced AI} \\
\textit{LG Electronics}\\
Santa Clara, CA, USA\\
Mohak.Shah@lge.com}}

\IEEEpubid{\makebox[\columnwidth]{978-1-7281-0858-2/19/\$31.00~\copyright2019 IEEE \hfill} \hspace{\columnsep}\makebox[\columnwidth]{ }}

\maketitle

\IEEEpubidadjcol

\newcommand{\aup}{{\it Auptimizer}\xspace}
\newcommand{\code}[1]{\texttt{#1}}

\newcommand{\ib}[1]{\textbf{\textit{#1}}}
\newcommand{\iuser}{{\it user}\xspace}
\newcommand{\ihyper}{{\it hyperparameter}\xspace}
\newcommand{\icode}{{\it code}\xspace}
\newcommand{\ijob}{{\it job}\xspace}
\newcommand{\iresearcher}{{\it researcher}\xspace}
\newcommand{\iexp}{{\it experiment}\xspace}

\newcommand{\jc}{\code{BasicConfig}\xspace}
\newcommand{\iproposer}{{\it Proposer}\xspace}

\renewcommand{\sectionautorefname}{Section}
\renewcommand{\subsectionautorefname}{Section} 
\renewcommand{\subsubsectionautorefname}{Section}
\newcommand{\lstlistingautorefname}{Code}
\renewcommand{\lstlistingname}{Code}
\newcommand{\algorithmautorefname}{Algorithm}

\begin{acronym}
  \acro{DS}{Data Science}
  \acro{ML}{Machine Learning}
  \acro{DL}{Deep Learning}
  \acro{HPC}{High Performance Computing}
  \acro{DNN}{Deep Neural Network}
  \acro{HPO}{Hyperparameter Optimization}
  \acro{RM}{{\it Resource Manager}}
  \acro{API}{Application Programming Interface}
  \acro{NAS}{Neural Architecture Search}
\end{acronym}

\begin{abstract}
  Tuning machine learning models at scale, especially finding the right hyperparameter values, can be difficult and
  time-consuming. In addition to the computational effort required, this process also requires some ancillary efforts
  including engineering tasks (e.g., job scheduling) as well as more mundane tasks (e.g., keeping track of the various
  parameters and associated results). We present \textbf{\emph{Auptimizer}}, a general \ac{HPO} framework to help data
  scientists speed up model tuning and bookkeeping. With \textbf{\emph{Auptimizer}}, users can use all available
  computing resources in distributed settings for model training. The user-friendly system design simplifies creating,
  controlling, and tracking of a typical machine learning project. The design also allows researchers to integrate new
  \ac{HPO} algorithms. To demonstrate its flexibility, we show how \textbf{\emph{Auptimizer}} integrates a few major
  \ac{HPO} techniques (from random search to neural architecture search). The code is available at
  \url{https://github.com/LGE-ARC-AdvancedAI/auptimizer}.
\end{abstract}


\section{Introduction}\label{sec:intro}

Designing a \ac{ML} framework for production faces challenges similar to those faced with Big Data. There is a large
{\bf volume} of models with a {\bf variety} of configurations and training them efficiently at scale with
reproducibility is critical to realizing their business {\bf value}. In this paper, we address one design aspect of the
\ac{ML} framework, namely the \ac{HPO} process, via a framework called \aup.

\subsection{\acl{HPO}}

\ac{ML} models are typically sensitive to the values of hyperparameters~\cite{vanrijn2018hyperparameter}. Different from
model parameters, these hyperparameters are values that control the model configuration or the training setup and thus
need to be set before training the model. Due to the lack of gradient information for these hyperparameters, tuning them
is often treated as a black-box optimization~\cite{golovin2017google}. As an alternative to manual selection (which is
usually based on modeler's expertise), researchers have proposed different methods to accelerate the tuning process
including Bayesian approaches \cite{snoek2012practical}, evolutionary algorithms~\cite{friedrichs2005evolutionary},
multi-armed bandits~\cite{falkner2018bohb}, and architecture search by learning~\cite{zoph2017neural}.

Tuning hyperparameters is often time-consuming especially when model training is computationally intensive
\cite{bergstra2012random}. Therefore, in practice, an \textbf{automated} \ac{HPO} solution is critically important for
machine learning. Both open-source solutions and commercial offerings are available.  However, as a rapidly
developing field, there are challenges when applying them under industry settings.

Specifically, no \ac{HPO} approach is objectively the best for all problems.  Most state-of-the-art open-source
solutions are backed by certain heuristics driven by research, e.g. \textsc{Spearmint}~\cite{snoek2012practical},
\textsc{HyperOpt}~\cite{bergstra2011algorithms,bergstra2013making}, and \textsc{HyperBand}~\cite{li2018hyperband}.
Sometimes users need to examine various options before settling on the most suitable approach.  But, transferring code
and results from one solution to another is difficult, given that no common \ac{API} is shared among them. This problem
has become more serious recently as the number of hyperparameters has increased dramatically with \ac{DNN}
algorithms, which are prone to human editing errors.

With the increased availability of large computing infrastructure, parallel and distributed training continue to become
affordable and commonplace. \ac{HPO} implementations typically do not allow users to fully benefit from such
high-performance computing environments. Furthermore, the complexity of the hyperparameter space and the long training
time make model tracking laborious and error-prone.

There have been efforts to automate this hyperparameter tuning process. \textit{Google Vizier}
\cite{golovin2017google} discussed the design and algorithms used for the Google Cloud Machine Learning \textit{HyperTune}
subsystem. Google AutoML, Amazon SageMaker, and SigOpt productize \ac{HPO} algorithms as commercial offerings. But no
customization is allowed on the algorithm or the infrastructure. Open-source projects like
\textit{Optunity}\cite{ClaesenMarc2014HtiP}, \textit{Tune}\cite{liaw2018tune} integrate different algorithms and have
built user-friendly \acp{API} for users. But, in all cases, adopting new algorithms or accommodating new computing
resources is still challenging. Also, hyperparameter tuning is often highly coupled with the respective framework and it
is difficult to use for other scenarios like fine-tuning an already trained model.

\subsection{Beyond \ac{HPO}}
Beyond the scope of \ac{HPO}, autoML tries to solve the \ac{ML} problem with minimal human intervention. Frameworks such
as ATM~\cite{swearingen2017atm}, auto-sklearn~\cite{feurer2015efficient}, auto-WEKA~\cite{kotthoff2016auto}, or
TuPAQ~\cite{sparks2015automating} based on the underlying \ac{ML} packages (sklearn, WEKA,
MLbase~\cite{pedregosa2011scikit,hall2009weka,kraska2013mlbase}) provide additional features such as model selection.
However, at the cost of simplifying the model engineering, practitioners also lose the ability to fully customize their
model or to use their existing model architectures.  Therefore, we focus on the scalability and automation of the
\ac{HPO} process and leave the model selection and feature engineering decisions for future work.

Our contributions are two-fold. We explicitly review the challenges of using \ac{HPO} in practice, and introduce an
open-source {\bf au}tomated o{\bf ptimizer} framework, \aup, to address these challenges by:

\begin{itemize}
  \item reducing the efforts to use and switch \ac{HPO} algorithms;
  \item providing scalability for cloud / on-premise resources;
  \item simplifying the process to integrate new \ac{HPO} algorithms and new resource schedulers;
  \item tracking results for reproducibility.
\end{itemize}

The paper is organized as follows. We summarize the state-of-the-art \ac{HPO} practices and outline the challenges in
\autoref{sec:background}. Next, we present the system design of \aup in \autoref{sec:design} and demonstrate its
usability in \autoref{sec:aup_use}. We further include the new development in the \ac{NAS} research in
\autoref{sec:nas}. We conclude our work with a discussion about the future roadmap of \aup.

%%%%%%%%%%%%%%%%%%%%%%%%%%%%%%%%%%%%%%%%%%%%%%%%%%%%%%%%%%%%%%%%%%%%%%%%%%%%%%%%%%%%%%%%%%%%%%%%%%%%%%%%%%%%%%%%%%%%%%%%
% Background
%%%%%%%%%%%%%%%%%%%%%%%%%%%%%%%%%%%%%%%%%%%%%%%%%%%%%%%%%%%%%%%%%%%%%%%%%%%%%%%%%%%%%%%%%%%%%%%%%%%%%%%%%%%%%%%%%%%%%%%%
\section{Background}\label{sec:background}

\subsection{Hyperparameter Optimization Research}\label{sec:hpo}

\ac{HPO} has become more relevant alongside the proliferation of \ac{ML} and data science applications.
\textsc{GridSearch} and manual search were favored in early studies due to their simplicity and
interpretability~\cite{larochelle2007an}. However, these approaches were quickly outpaced by others due to the
\textit{curse of dimensionality}~\cite{bergstra2012random}.

Different hyperparameters are not independent. Instead, their values are intertwined. Advanced algorithms take advantage
of this internal constraints to balance exploration and exploitation of the parameter spaces to offer a better solution.
Modeled by a Gaussian process, Bayesian Optimization tries to maximize the expected model improvement and works well
for low-dimensional, numerical problems (e.g., \textsc{Spearmint}~\cite{snoek2012practical}). Sequential Model-based
Global Optimization with tree-based method \cite{bergstra2011algorithms} has been shown to have better performance in
high-dimensional, structured model space \cite{eggensperger2013towards}. Recently, optimizing
\ac{DNN} models using reinforcement learning has become a mainstream topic  under the rubric of \ac{NAS}
\cite{zoph2017neural}.

Besides learning the structure of hyperparameters, optimizing the training budget using multi-armed bandit strategy also
shows promising results for \ac{DNN} models (e.g., \textsc{HyperBand} \cite{li2018hyperband}).  Further combining with
Bayesian optimization, \textsc{BOHB}~\cite{falkner2018bohb} improves the tuning process leveraging on the benefits of both
approaches. Despite its simplicity, \textsc{RandomSearch} \cite{bergstra2012random} is still efficient and is commonly
used as a benchmark against other more advanced algorithms \cite{dutta2018effective,li2018hyperband}. 
Given the variety of \ac{ML} problems, there is no conclusive preference of the best \ac{HPO} to use.
% The variety of HPO choices and no no-deterministic benefits create difficulties for practitioners to accept them for applications.

Regardless of their variety, all \ac{HPO} algorithms share the same workflow, which breaks down into
four steps: 
\begin{enumerate}
  \item initialize search space and configuration; \label{step1}
  \item propose values for hyperparameters; \label{step2}
  \item train model and update result; \label{step3}
  \item repeat step \ref{step2} and \ref{step3}. \label{step4}
\end{enumerate}
Most of above research focus on the Step~\ref{step2}, and the engineering-oriented projects discussed next are focused
more on streamlining the Step~\ref{step4} while supporting a limited number of algorithms. Our proposed framework, \aup,
helps automate the entire process.

\subsection{Hyperparameter Optimization Practice}\label{sec:practice}

Most of the above-mentioned algorithms and many others have released their source codes to the research community, e.g.,
\textsc{Spearmint}\footnote{\url{https://github.com/JasperSnoek/spearmint}},
\textsc{HyperOpt}\footnote{\url{https://github.com/hyperopt/hyperopt}}, and
\textsc{HyperBand}\footnote{\url{https://github.com/zygmuntz/hyperband}}. These solutions are originally designed for
research. Therefore they are hard to extend or integrate with other algorithms. Moreover, neither common code structure
nor common \acp{API} exist for interoperability. Thus, it is challenging for users to adopt them without changing their
existing code and to switch amongst these alternatives without significant changes to their code base.

The efforts to consolidate the interfaces at a system level are also available. Google's
\textit{Vizier}\footnote{Unofficial source at \url{https://github.com/tobegit3hub/advisor}}~\cite{golovin2017google} is a
design
used within Google for their Cloud Machine Learning \textit{HyperTune} subsystem. And Google AutoML, Amazon SageMaker,
and SigOpt productize \ac{HPO} algorithms as commercial offerings. But these approaches typically fail to provide the
extensibility for users to integrate their specific \ac{HPO} algorithms or the scalability to utilize a large pool of
computing resources on-premise.

Open-sourcing helps the extensibility, however, the existing packages often fall short on challenges to practical
use. Projects like \textsc{Optunity}\footnote{\url{https://github.com/claesenm/optunity}} or
\textsc{Chocolate}\footnote{\url{https://github.com/AIworx-Labs/chocolate}} integrated a few different \ac{HPO}
algorithms for users, and new ones can be easily integrated under their consistent \acp{API}. However, the process is
sequential, therefore they do not support training models parallelly at scale. On the other side,
\textsc{Dask-ML}~\cite{rocklin2015dask} provides an easy-to-switch backend for computing resources, but lacks supports
for customizing and extending \ac{HPO} algorithms.

\textsc{Tune}~\cite{liaw2018tune}, centering on scalable hyperparameter search, is undergoing rapid development. It
supports scalability on different architectures, and also supports two categories of \ac{HPO} strategies: trial
schedulers (e.g. \textsc{HyperBand}) and search algorithms (e.g. \textsc{HyperOpt}). However, it lacks usability in
practice. First, the users' training script needs modification to align with \textsc{Tune}'s \ac{API}. This approach
occasionally results in excessive re-engineering on source code and it hinders users from debugging their training code.
Second, different search algorithms require different configurations, which makes it harder for users to switch among
different \ac{HPO} strategies. Third, \textsc{Tune} relies on the autoscaling function provided by the \textsc{Ray}
project for computing resource allocation, which currently cannot support a team environment where more advanced job
scheduler is already in place.

There is no universal \ac{HPO} algorithm having the best performance over all problems.   Thus, trying different ones is
necessary to reveal the best results and business value.  However, a high adoption cost commonly prevents user from
trying different algorithms.  We summarize the common factors that limit the current \ac{HPO} toolboxes as flexibility,
usability, scalability, and extensibility:

\begin{itemize}
  \item Flexibility.  It is challenging to switch between \ac{HPO} algorithms, as the interfaces are dramatically
  different.
  \item Usability. It is time-consuming to integrate an existing \ac{ML} project into an \ac{HPO} package. Often, users
  need to rewrite their code for a specific \ac{HPO} toolbox, and resulting script cannot be used anywhere else.
  \item Scalability. The integration with large-scale computational resources is missing and it is typically hard to
  scale the toolbox to a multi-node environment.
  \item Extensibility. It is challenging to introduce a new algorithm into the existing libraries as these libraries are
  tightly coupled with the implemented algorithms.  
\end{itemize}

We summarize the comparison of representative \ac{HPO} solutions based on the above criteria in \autoref{tab:toolbox}.
Based on our experience in developing an in-house solution, we release an \ac{HPO} framework, \aup, to
mitigate the above-mentioned challenges.

\begin{table*}
  \caption{Comparison of \ac{HPO} toolboxes.}\label{tab:toolbox}
  \centering
  \begin{tabular}{|l|l|l|l|l|l|l|}
    \hline
    Criteria        & \textsc{HyperOpt} & SageMaker & \textsc{Optunity} & \textsc{Dask-ML}  & \textsc{Tune} & \aup \\
    \hline
    Open source     & Yes               & No        & Yes               & Yes               & Yes           & Yes \\
    Flexibility (No. of \ac{HPO} algorithms)    & 2                 & Bayesian  & 7                 & 2                 & 4, 8          & 9   \\
    Usability (Format of training code)      & Function          & Rewrite   & Function          & Rewrite           &
    Function      & Script \\
    Scalability     & Manual            & Cloud     & No                & Yes               & Yes           & Yes \\
    Extensibility (Manual to add new \ac{HPO} algorithms)  & N.A.              & N.A.      & Yes               & Hard              & Yes           & Yes \\
    \hline
  \end{tabular}
\end{table*}

\subsection{Definitions}\label{sec:definitions}
In this paper, we use the following terminology to describe the system design and \aup use cases.

For data science applications, data scientists (\ib{user}s) solve given data mining problems with specified \ac{ML}
models. A script (\ib{code}) is written and some \ib{hyperparameter}s are commonly identified to be explored during the
model training. Typically, the \iuser carries out an \ib{experiment} to examine a range of \ihyper combinations and
measures the performance (e.g., accuracy) of the model on a hold-out dataset, for example, the number of neighbors in a
K-Nearest-Neighbor model, or the learning rate in a deep learning model. Each individual training process for a given
\ihyper set is called a \ib{job}. After all \ijob{s} are finished, the \iuser retrieves the best model from the
training history for further analysis or application.

For \ac{ML} \ib{researcher}s in the \ac{HPO} field, the use case is different. \textit{Researcher}s
focus on developing the \ib{algorithm} to find the best hyperparameters. Thus, an easy framework to facilitate their
algorithm implementation and to benchmark their results against the state-of-the-art algorithms is important.
%\citealt{eggensperger2013towards} have provided an \ac{HPO} benchmark, and we focus on the framework of development and
%deployment.

%We discuss the design of \aup and explain how it helps the \ac{ML} research community. In \autoref{sec:aup_use} we
%discuss \aup usability in practice, and it is relevance to the data science community.

%%%%%%%%%%%%%%%%%%%%%%%%%%%%%%%%%%%%%%%%%%%%%%%%%%%%%%%%%%%%%%%%%%%%%%%%%%%%%%%%%%%%%%%%%%%%%%%%%%%%%%%%%%%%%%%%%%%%%%%
% Design section
%%%%%%%%%%%%%%%%%%%%%%%%%%%%%%%%%%%%%%%%%%%%%%%%%%%%%%%%%%%%%%%%%%%%%%%%%%%%%%%%%%%%%%%%%%%%%%%%%%%%%%%%%%%%%%%%%%%%%%%
\section{Design}\label{sec:design}

\aup is designed primarily as a tool for \iuser.  It removes the burden of drastically changing users' existing code,
which is a key hurdle in the \ac{HPO} adoption process.  It only requires the \iuser to {\bf add} a few lines in the
code, and guides \iuser{s} to setup all other experiment-related configurations. Therefore, the \iuser can easily
switch among different \ac{HPO} algorithms and computing resources without rewriting their training script.

\begin{figure}
  \centering
  \includegraphics[width=\linewidth]{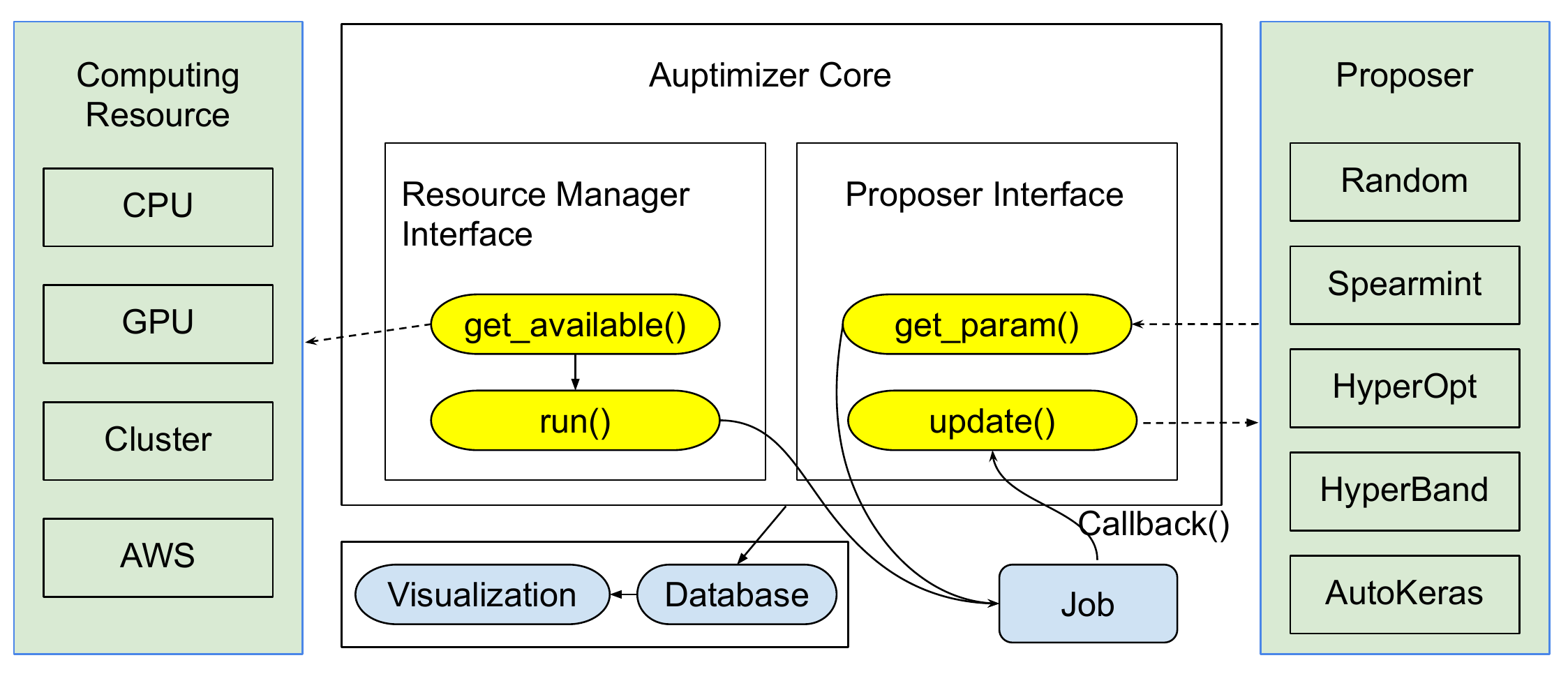}
  \caption{System Design}
  \label{fig:system}
\end{figure}

\aup is also designed to support researchers and developers to easily extend the framework to other \ac{HPO} algorithms
and computing resources. We highlight the abstraction of the \aup design in \autoref{fig:system}.  Both resources and
proposers communicate with \aup via the designated interfaces.  We implemented a few open-source \ac{HPO} solutions to
demonstrate the consistency of the \ac{API} definitions. Meanwhile, the user's training script is executed as a \ijob,
in which the scores are automatically updated for proposer without user's intervention.

The \aup framework abstracts the \ac{HPO} workflow of an \iexp as shown in \autoref{alg:hpo}. Once an \iexp is defined
and initialized, \aup continuously checks for available resources (\code{get\_available()}) and new \ihyper proposals
(\code{get\_param()}) and then runs new \ijob{s} to search for the best model. Once a job is finished, \aup
automatically starts \code{update()}, a function that records the results asynchronously using a callback mechanism.

\begin{algorithm}
  \caption{\aup Internal Workflow}\label{alg:hpo}
  \begin{algorithmic}
    \REQUIRE experiment.json; env.ini; code\_path
    \STATE aup.Experiment(experiment.json, env.ini, code\_path)
    \WHILE {not	proposer.finished()} 
    \STATE resource $\leftarrow$ resource\_manager.get\_available() 
    \IF{not resource}
    \STATE sleep 
    \COMMENT{wait for available resource}
    \ENDIF
    \STATE hyperparameters $\leftarrow$ proposer.get\_param()
    \STATE Job $\leftarrow$	aup.run(hyperparameters, resource) 
    \IF{Job.callback()} 
    \STATE proposer.update()
    \ENDIF
    \ENDWHILE
    \STATE aup.finish()
    \COMMENT{wait for	unfinished jobs}
  \end{algorithmic}
\end{algorithm}

In the following sections, we discuss the two key components - \acl{RM} and \iproposer - along with auxiliary components
- Tracking and Visualization - in detail.  Researchers and developers will find that these abstractions can help them to easily
extend \aup with new \ac{HPO} algorithms and adapt it to their own computing environments.

\subsection{\iproposer}\label{sec:proposer}

\iproposer controls how \aup interacts with \ac{HPO} algorithms for recommending new hyperparameter values.  The
\iproposer interface reduces the effort to implement an \ac{HPO} algorithm by defining two functions:
\code{get\_param()} to return the new hyperparameter values, and \code{update()} to update the history.  In the open
source release, we integrate a few well-known solutions, such as \textsc{Spearmint}~\cite{snoek2012practical},
\textsc{HyperOpt}~\cite{bergstra2011algorithms,bergstra2013making}, \textsc{HyperBand}~\cite{li2018hyperband},
\textsc{BOHB}~\cite{falkner2018bohb} along with simple random search and grid search.  Moreover, we also demonstrate its
usability to a state-of-the-art \ac{NAS} approaches such as \textsc{EAS}~\cite{cai2018efficient} and AutoKeras~\cite{jin2018efficient}.  

Despite the inherently different nature of these algorithms, \aup interacts with them only through the two interfaces
described above and keeps other irrelevant components away from users and researchers.  When implementing other
open-source solutions, we found that at most one source file needs to be changed or added, and the remaining source code
can be reused for the integration.  As an example, to integrate \textsc{BOHB}, we wrote only 138 lines of code and
reused the existing 4305 lines of codes\footnote{\url{https://github.com/automl/HpBandSter} with commit \em{841db4b}.},
which demonstrates the power of \aup's {\bf extensibility}.

\subsubsection{\code{get\_param()}}\label{sec:get}
function is a wrapper for the underlying \ac{HPO} implementations. It queries new values of
\ihyper{s} and package them into a \code{BasicConfig} object to be used for \icode execution.

The newly created \jc contains all \ihyper values in a dictionary for a \ijob to run with. Additional information can be
added for \ac{HPO} algorithms to use without interfering with job execution.  For instance, the value of the job ID is
used in the \textsc{HyperBand} implementation to track previous results and to resume training when
necessary. This \code{BasicConfig} is then passed to the resource manager for job execution (see discussion in
\autoref{sec:rm}). \textit{User}s only need to change their \icode to read the \jc as an input file. To further reduce the
burden on the \iuser end, we provide \code{load()} and \code{save()} methods in \code{BasicConfig} to simplify the
adoption of \aup (see example in \autoref{lst:main}).

An example of the \jc file generated by \aup at runtime is illustrated in \autoref{lst:job_config}. It contains two
variables (\code{x,y}) along with additional variables when necessary (e.g., \code{job\_id}). This generated \code{JSON} file will
be passed to the \icode automatically by \acl{RM} during model training.

\begin{lstlisting}[caption=Job Configuration File,label={lst:job_config},float]
  {"x": -5.0, "y": 5.0, "job_id": 0}
\end{lstlisting}

All configurations used for model training are saved, and user can easily reuse them together with their code without
any modification.  This enables users to verify or finetune their model after \ac{HPO}.

\subsubsection{\code{update()}}
function collects results back from \ijob{s}, updates the tuning history, and also registers the
results for record tracking (see \autoref{sec:tracking}).

For simple algorithms (e.g., \textsc{RandomSearch}), no history is needed. However, advanced algorithms (e.g., Bayesian
Optimization) need to match the resulting scores with the specific input hyperparameters. \aup takes care of this
matching by automatically mapping the result back to its \jc and thus, \ac{HPO} algorithms can directly restore the
\ihyper values used in a specific \ijob.  Auxiliary values (e.g. \code{job\_id}), are tracked and can be customized for
other usage, such as to save and retrieve models for further finetuning\footnote{Users need to write their own function
to restore model based on the input ID.}.

\begin{figure*}[htbp]
  \begin{center}
    \includegraphics[width=0.91\textwidth,trim={0 1cm 0 0},clip]{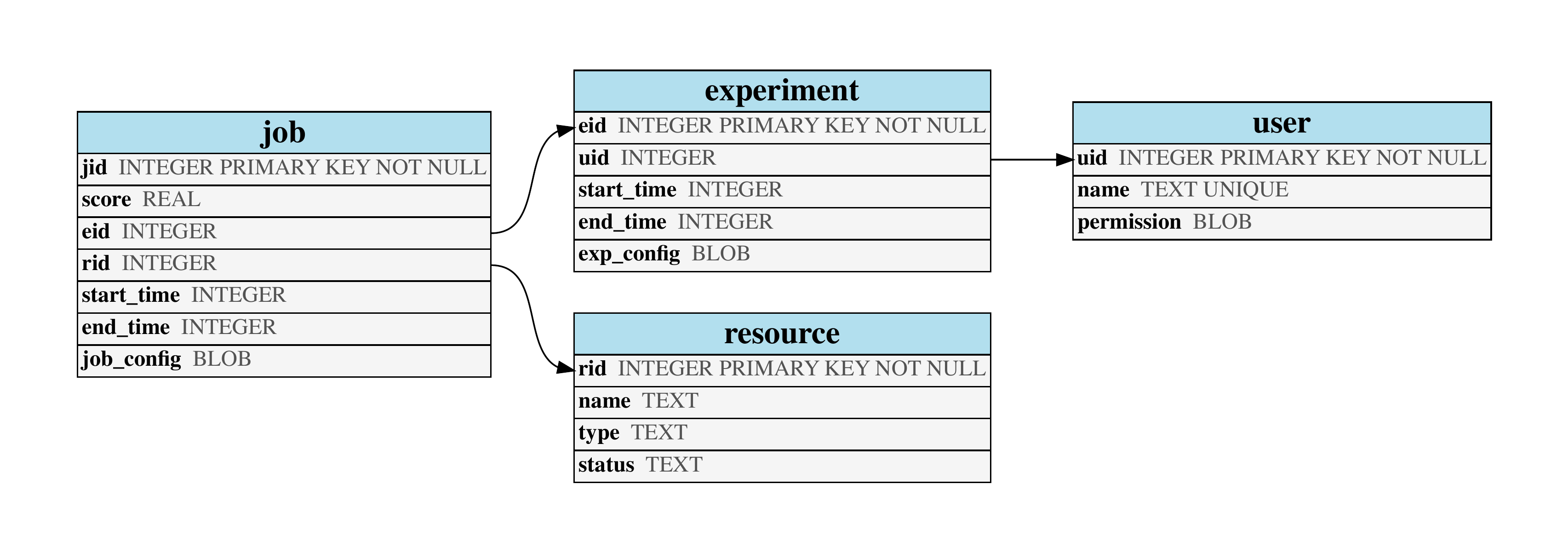}
    \caption{Database Schema}
    \label{fig:database}
  \end{center}
\end{figure*}

\subsection{\acl{RM}}\label{sec:rm}

\ac{RM} is another cornerstone in the \aup framework. It connects computing resources to model training automatically
thus allowing \icode{s} to run on resources based on their availability. It also sets a callback mechanism to trigger
the \code{update()} function when a \ijob is finished.

A key challenge of usability in \ac{HPO} implementations is the communication between \ijob{s} and the heterogeneous
computer resources that \ijob{s} run on.  The existing open-sourced projects, e.g., \textsc{HyperOpt}, \textsc{Tune},
require to call the code directly to get return values.  And commercial services such as SageMaker requires its
customer's code to be encapsulated in a docker image.  SigOpt provides \ac{API} calls for communication but it leaves it
to the \iuser{s} to do resource allocation and code execution.  All these solutions are challenging for \iuser{s} to use
the \ac{HPO} at scale.  In comparison, the \aup framework puts the user-friendliness as its priority and removes this
burden.

General resource management and job scheduling tools, e.g., Slurm \cite{yoo2003slurm} or TORQUE \cite{staples2006torque}
are not designed for \ac{HPO} applications. Using those tools, jobs are submitted in advance and wait for the available
resources to be executed on. In the \ac{HPO} setting, the configurations of hyperparameters are typically determined
based on the history of model scores and it results in a difficulty to start \ijob{s} spontaneously. Without \aup,
\iuser{s} need to either allocate all resources at once or to write their corresponding interface to start new \ijob{s}
on the fly. However, in our workflow, we rely on the flexibility of cloud services (i.e. AWS) to scale out.
For Slurm and other tools, we are open to community support.

The \ac{RM} interface makes it simple to extend \aup to scalable computing environments.  Developers only need to
interact with \code{get\_available()} and \code{run()}, which queries available resources and allocate correspondingly
for job execution.  As users, they only need to specify the resources to be used in the experiment configurations. Also
integrating with advanced scheduling tools, \aup can further help \iuser{s} to  schedule jobs efficiently in a
multi-tenant environment with better resource allocation. 

\subsubsection{\code{get\_available()}} function serves as the interface between \aup and typical resource management and job
scheduling tools. In the current implementation, it queries a persistent database for available resources that the
\iuser specified. If the requested resource is available, then it will be taken by \aup for job execution. Otherwise,
the system will wait until resources are free (see \autoref{alg:hpo}).

The interface \code{get\_available()} is also compatible with existing resource management tools. For instance, we used
\code{boto 3}\footnote{\url{https://github.com/boto/boto3}.} to spawn new EC2 instances on the AWS.

\subsubsection{\code{run()}}

The \aup \ac{RM} component relies on the callback design to solve the scheduling wrapped in the \code{run()} function.
Specifically, \code{run()} interface executes the user-provided code in a \code{Job} object. The \code{Job} object first
sets up the running environment based on the available resources. For instance, it assigns \code{CUDA\_VISIBLE\_DEVICES}
for GPU allocation. Then it executes the user-provided \icode with the newly proposed hyperparameter values (see
\autoref{sec:get}). Once a job is finished, it triggers a \code{callback()} function to \code{update()} the result in
\aup.  It also allows additional information to be passed to \iproposer as an arbitrary string when returned from users'
code.

In the current version of \aup, we demonstrate its usability across different computing resources, such as CPUs, GPUs,
multiple nodes, and AWS EC2 instances. We require that the users' code executes successfully on the targeted resources
to avoid potential environment issues. In this initial release, we use a SQLite database to keep track of available
resources and all jobs are running locally. Both the hyperparameter configuration and the results are communicated by
the standard IO protocol. 

\subsection{Experiment Tracking and Visualization}\label{sec:tracking}

Experiment tracking provides a foundation of reproducibility in a data science project. In \aup, all the experiment
history is tracked in the user-specified database. The data schema is illustrated in \autoref{fig:database}.

\code{Experiment} table plays the central role to track the overall progress. It contains experiment ID, user ID, and
start and end time of an \iexp. Beside them, the \code{exp\_config} specifies the scope of the \iexp (see
\autoref{sec:ec} for detailed discussion) The tables of \code{User} and \code{Resource} are for user control and resource
management. The \code{Job} table tracks the runtime status and result of each \ijob. Since \aup automatically checks in
its training process in \iexp{s}, \iuser{s} are alleviated from the worry of losing reproducibility.

The \aup framework also provides a basic tool to visualize the results from history (see \autoref{sec:demo}).  In
addition, users are able to directly access the results stored in the database for further analysis.

%%%%%%%%%%%%%%%%%%%%%%%%%%%%%%%%%%%%%%%%%%%%%%%%%%%%%%%%%%%%%%%%%%%%%%%%%%%%%%%%%%%%%%%%%%%%%%%%%%%%%%%%%%%%%%%%%%%%%%%%
% How to
%%%%%%%%%%%%%%%%%%%%%%%%%%%%%%%%%%%%%%%%%%%%%%%%%%%%%%%%%%%%%%%%%%%%%%%%%%%%%%%%%%%%%%%%%%%%%%%%%%%%%%%%%%%%%%%%%%%%%%%%
\section{Using \aup}\label{sec:aup_use}

In this section, we demonstrate the key features of using \aup in practice, by the simple and commonly used \ac{DNN}
model for the MNIST
dataset\footnote{\url{https://github.com/aymericdamien/TensorFlow-Examples/blob/master/examples/3_NeuralNetworks/convolutional_network.py}
with commit \it{971c96b}.}\cite{lecun1998gradient}. The \ac{DNN} model contains two convolution layers and two fully
connected layers. Adam optimizer is used for training \cite{kingma2014adam} with a global dropout ratio for
regularization \cite{srivastava2014dropout}.  Also, for demonstration purpose, we only search for the best accuracy on
the test dataset without distinguishing it from the validation dataset.

\subsection{\aup Workflow}\label{sec:aup_workflow}

In this section, we illustrate the basic workflow to adopt \aup.

First, we need to set up the \aup by filling in the basic information for the computing environment.  \aup has a
user-friendly interactive guide that can be invoked by \code{python -m aup.setup}. It will setup the \aup for the first
time with information about the computing environment and the database.

Next, we need to identify the key \ihyper{s}. In this experiment, we will explore five hyperparameters, numbers of
filters in the first two convolution layers (conv1, conv2), the dropout ratio, the number of neurons of the first
fully-connected layer, and the learning rate.  Also, we use \code{n\_iterations} to adjust number of epochs in training,
which is useful for \textsc{HyperBand} and \textsc{BOHB}. 

After that, we need to write down the experiment configuration correspondingly, which we explain in \autoref{sec:ec}.
And the training script is modified accordingly in \autoref{sec:code}.  

\subsection{Experiment Configuration}\label{sec:ec}

\aup also provides a command-line tool to initiate the file as \code{python -m aup.init}. Experiment Configuration
controls the search space and the choice of \ac{HPO} algorithm of an \iexp. In \autoref{lst:exp_config}, we
illustrate the configuration for random search for the Rosenbrock function \cite{rosenbrock1960an}.

\autoref{lst:exp_config} shows that the configuration is simple and straightforward. The \code{n\_samples} specifies how
many \ijob{s} a \iuser wants to run for the \ac{HPO} process and \code{n\_parallel} \ijob{s} can be executed at the same
time on the CPU \code{resource}. The \ihyper space is defined in \code{parameter\_config}, each \ihyper is a float
ranging from $-5$ to $10$. Those \ihyper{s} will be assigned by \aup into a \jc (e.g., \autoref{lst:job_config}) and
will be accessed from the \iuser{'s} \icode directly (see \autoref{lst:main}).

Occasionally additional information can still be required for different \ac{HPO} algorithms, e.g., using
\code{"engine":"tpe"} to instruct \textsc{HyperOpt} to use \textsc{TPE} as the backend engine for \ac{HPO}. But overall,
the change in experiment configuration is significantly reduced in contrast to using  different open-source
implementations of \ac{HPO} algorithms. And most importantly, there is no need to change the user's code for different
algorithms.

\aup can also guide through the process of selecting \ac{HPO} algorithm and defining hyperparameter specifications
interactively.  One generated example configuration file for random search is shown in \autoref{lst:exp_config}.  Users
enjoy {\bf flexibility} and {\bf scalability} by simply changing the proposer name or the parallel number.

\begin{lstlisting}[caption=Experiment Configuration File,label={lst:exp_config},float]
  {
    "proposer": "random",
    "script": "mnist.py",
    "resource": "gpu",
    "n_parallel": 2,
    "target": "min",
    "parameter_config": 
    [
      {"name": "conv1", "range": [20, 50],  "type": "int"},
      {"name": "dropout", "range": [0.5, 0.9], "type": "float"},
      ...
    ],
    "n_samples": 100
  }
\end{lstlisting}

\subsection{Code Update}\label{sec:code}

To run jobs automatically in \aup, we need to modify the source code correspondingly.  Comparing to other \ac{HPO}
tools, the modifications are significantly reduced, and the resulting code can still be run independently without \aup.
Generally, there are four items: 
\begin{itemize}
  \item change the code to self-executable, 
  \item parse input hyperparameters,
  \item use them for model training, 
  \item report back the result. 
\end{itemize}

We highlight the changes in \autoref{lst:main} and the steps are explained here:

\begin{enumerate}
  \item Line 1: add the shebang line to make the code self-executable.
  \item Line 2-3: import \code{sys, aup} to parse hyperparameters and return values to \aup.
  \item Line 4-5: modify original training function by replacing variables with hyperparameters in \code{config}.
  \item Line 6-7: main function.  It parses \jc by the input file.
  \item Line 8-9: original training script.  It trains for \code{config['n\_iterations']} epochs and computes the test
  accuracy.
  \item Line 10: return score to \aup.
\end{enumerate}

\begin{lstlisting}[caption=Example Code,label={lst:main},numbers=left,float,xleftmargin=2em,
  linebackgroundcolor={
  \ifnum\value{lstnumber}=1\color{yellow}\fi
  \ifnum\value{lstnumber}=3\color{yellow}\fi
  \ifnum\value{lstnumber}=7\color{yellow}\fi
  \ifnum\value{lstnumber}=10\color{yellow}\fi}]
  #!/usr/bin/env python
  import sys
  from aup import BasicConfig, print_result
  config = dict(conv1=32, conv2=64, fc1=1024, learning_rate=0.001, dropout=0.1, data_dir="input_data", n_iterations=10)
  # Training code with variables replaced as hyperparameters
  if __name__ == "__main__":
    config = BasicConfig(**config).load(sys.argv[1]) if len(sys.argv) > 2
    accuracy = ... # training code returns test accuracy
    print_result(accuracy) 
\end{lstlisting}

As demonstrated above, minimal changes are required to fully adopt the \aup framework into practice.  More importantly,
the {\bf usability} of the code remains, and users can reuse the exact same script for other purposes (training from
scratch, finetuning) by providing the hyperparameters as input.  Moreover, \aup does not restrict users on any specific language
or framework.  For instance, a \textsc{MatLab} user can also use \aup to tune their hyperparameters once they parse and
return result in their code correspondingly.

%%%%%%%%%%%%%%%%%%%%%%%%%%%%%%%%%%%%%%%%%%%%%%%%%%%%%%%%%%%%%%%%%%%%%%%%%%%%%%%%%%%%%%%%%%%%%%%%%%%%%%%%%%%%%%%%%%%%%%%%
% Result / Demo
%%%%%%%%%%%%%%%%%%%%%%%%%%%%%%%%%%%%%%%%%%%%%%%%%%%%%%%%%%%%%%%%%%%%%%%%%%%%%%%%%%%%%%%%%%%%%%%%%%%%%%%%%%%%%%%%%%%%%%%%
\subsection{Experiments}\label{sec:demo}

After defining the experiment and modifying code, \iuser{s} are ready to run the experiment by simply entering
\lstinline{python -m aup experiment.json}. More importantly, \iuser{s} can easily switch among \ac{HPO} algorithms by
updating the configuration, or retrieve and store results in the database.

We allocate roughly the same number of total training epochs for each \ac{HPO} algorithm.  For random, \textsc{Spearmint},
\textsc{HyperOpt}, each hyperparameter configuration is trained for 10 epochs with 100 different configurations. Whereas
for grid search, we assign the grid with 3 values for all hyperparameters, except the learning rate which is chosen
from $0.001,0.01$, resulting in 162 configurations.  For \textsc{HyperBand} and \textsc{BOHB}, we allocate a total budget
of 1000 epochs approximately along with 100 configurations to be explored.  We also enforce the minimum number of epochs
to be 1 with no upper limit.

\begin{figure}
  \centering
  \includegraphics[width=0.9\linewidth]{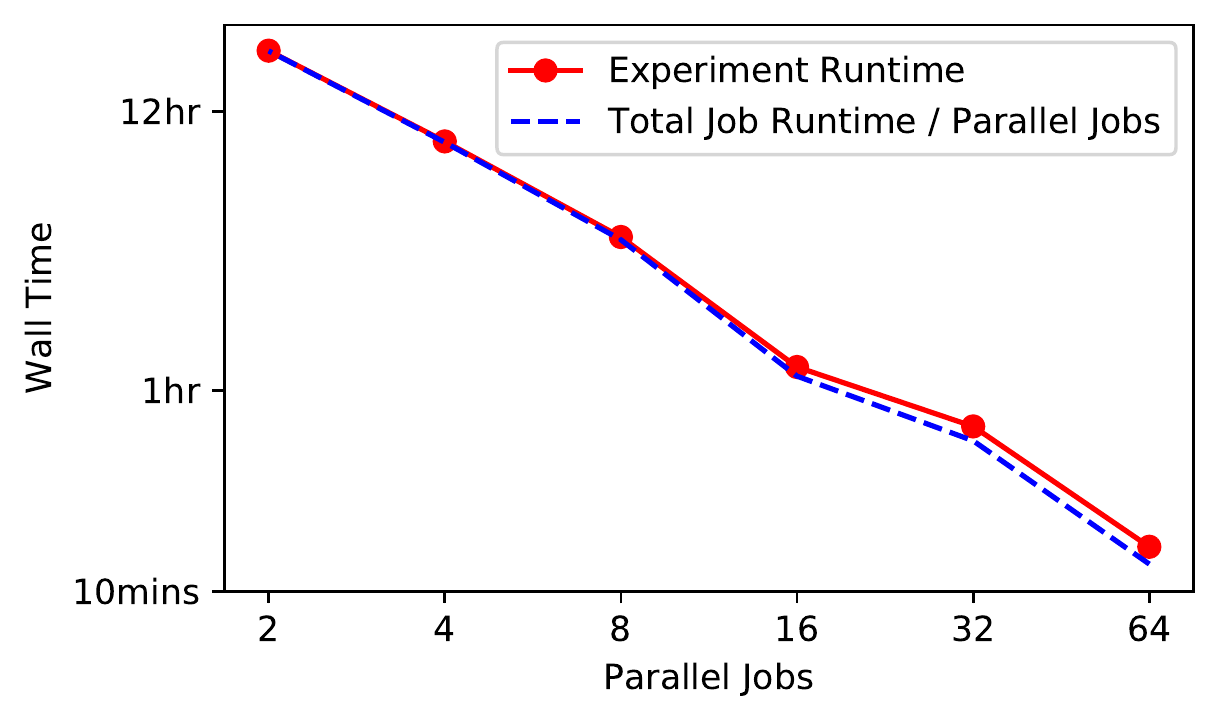}
  \caption{\aup scalability on AWS}
  \label{fig:sfig1}
\end{figure}

In \autoref{fig:sfig1}, we examine the scalability of \aup by comparing the overall experiment time with the total time
used by all jobs divided by the number of computing resources.  The experiment searched for 128 configurations with up
to 64 AWS EC2 instances.  Because training time varies due to the changing model complexity (i.e. number of filters and
neurons), we fixed the random seed, such that all experiments explored the same configurations. On average, each job ran
5 minutes on a t2.medium instance with 4 vCPUs.  Clearly, the training time dominates the runtime, whereas the
communication and the \ac{HPO} algorithm (random) take marginal time in total. The break from linearity is caused by two
issues.  First, the total time of an experiment is driven by the last job.  Because different jobs have different
training times, the gap between experiment time and the total time used by jobs becomes larger when using more parallel
machines.  Second, the performance fluctuation of the EC2 machines is the main reason for the nonlinearity in the
scaling relation and it is not controllable by \aup. More importantly, typical model training time is much longer than 5 minutes, which makes the
additional cost by \aup negligible.

\begin{figure}
  \centering
  \includegraphics[width=.9\linewidth]{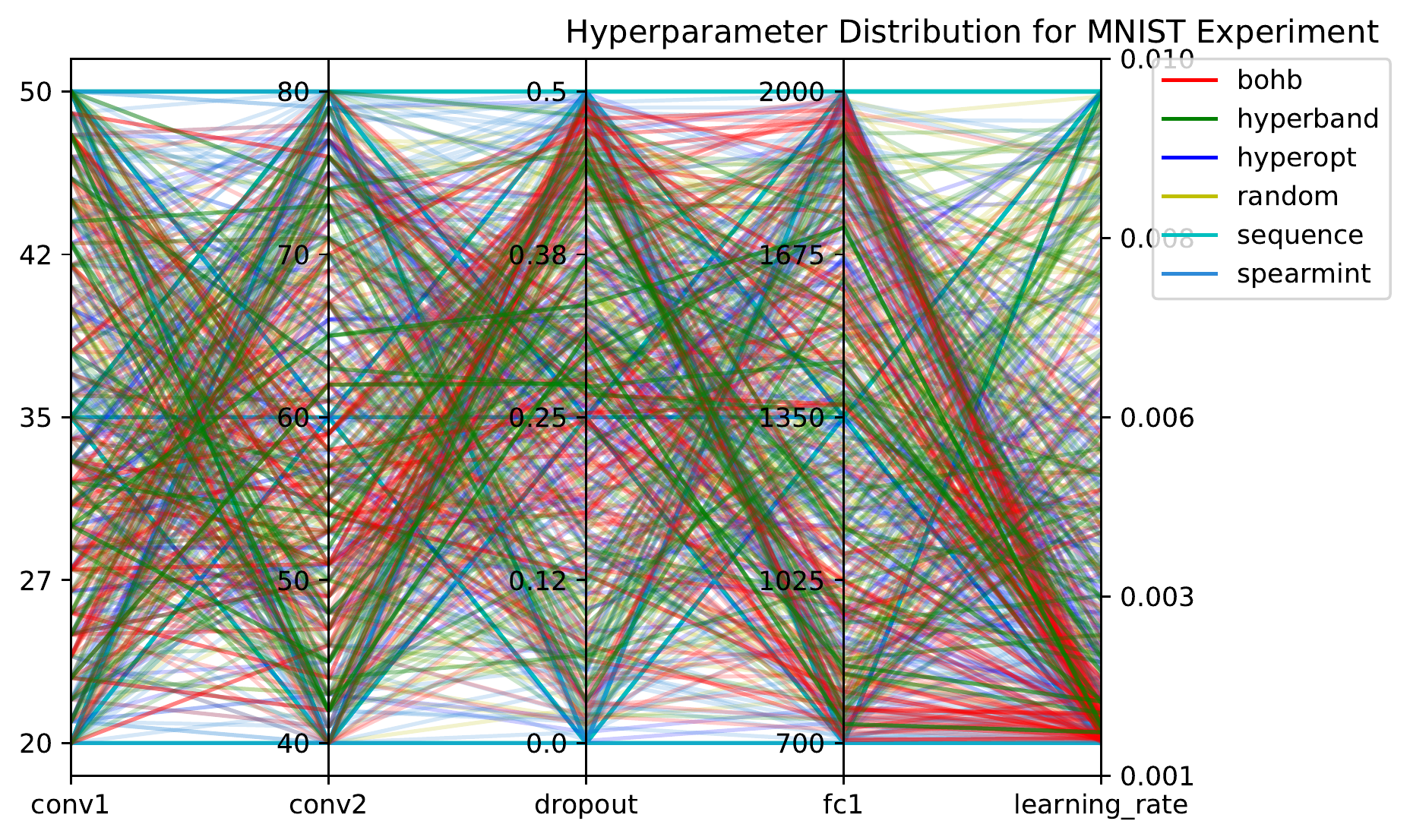}
  \caption{Hyperparameter Distribution from Different \ac{HPO} Algorithms}
  \label{fig:sfig3}
\end{figure}

We illustrate all hyperparameter combinations from different \ac{HPO} algorithms in \autoref{fig:sfig3}.  It shows that
different \ac{HPO} algorithms have searched for different paths in the hyperparameter space. Choosing the optimal
\ac{HPO} algorithm is a challenge and exploring them easily is important in practice.  Among different approaches, we
only need to change the name of algorithms, which significantly reduce the engineering work at the code level.  Users
can also easily scale the experiment to run in parallel by specifying the \textit{n\_parallel} value. Researchers can
use it as benchmark suite when they have a new algorithm to test against.

\begin{figure*}
  \centering
  \includegraphics[width=.97\linewidth]{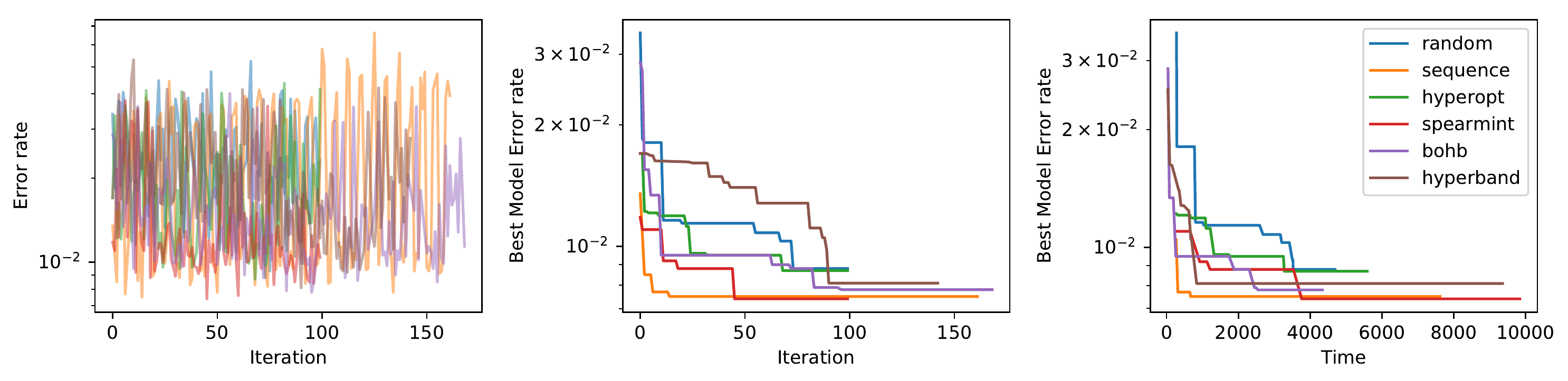}
  \caption{Performance of Different \ac{HPO} Algorithms}
  \label{fig:sfig2}
\end{figure*}

In \autoref{fig:sfig2}, we show the performance of different \ac{HPO} algorithm with \code{n\_parallel=8}. We want to
emphasize that the purpose is to demonstrate the usability of \aup rather than to benchmark different \ac{HPO} algorithms.
By changing the algorithm names, we can easily run different strategies to tune a given model. Albeit the error rate
reported here is not representative as no validation set is used. 
%and proposed models did not overfit.  
We can still
confirm a few characteristics of the \ac{HPO} algorithms. For example, \textsc{Spearmint} generally find good models at
the cost that most models are complex models and result in longer training times.   And as expected, \textsc{BOHB} and
\textsc{HyperBand} are more resource efficient in finding good models. Grid search explored the complicated
model at the early stage, which lead to an overall good performance, but in practice, when the reasonable range is not
available or the dimensionality is high, it often does not work well.

%%%%%%%%%%%%%%%%%%%%%%%%%%%%%%%%%%%%%%%%%%%%%%%%%%%%%%%%%%%%%%%%%%%%%%%%%%%%%%%%%%%%%%%%%%%%%%%%%%%%%%%%%%%%%%%%%%%%%%%%
% NAS
%%%%%%%%%%%%%%%%%%%%%%%%%%%%%%%%%%%%%%%%%%%%%%%%%%%%%%%%%%%%%%%%%%%%%%%%%%%%%%%%%%%%%%%%%%%%%%%%%%%%%%%%%%%%%%%%%%%%%%%%
\section{Neural Architecture Search}\label{sec:nas}
Though neural networks have become ubiquitous for various AI tasks there is still a lot of expert knowledge needed for
designing architectures. As a result, recently gradient-based architecture search has become very popular. A seminal
paper describing the research is \cite{zoph2017neural}, which uses a recurrent-network-based ``controller" to generate
strings of ``child nets". These child nets are also neural networks, whose architectures are specified by a string
variable and are each trained to convergence. The controller then uses the accuracy of the child nets as a reward signal to
compute the policy gradient. Progressively the controller will give higher probabilities to architectures with higher
accuracy and improves its search over time, learning architectures which would progressively improve accuracy. The same
paradigm can be easily adapted and extended using \aup where the controller can be abstracted into \iproposer, allowing
users to both improve and develop \ac{NAS} algorithms in a scalable and automated environment.

Since \cite{zoph2017neural}, further improvements have been suggested. However, the essential technique and approach
remains the same. In \cite{pham2018efficient}, the authors show how to improve the efficiency of \ac{NAS} by forcing all
child models to share weights, which allows child networks to train efficiently to convergence without starting from
scratch every time. 

The technique works by incorporating transfer learning between child models which substantially reduces the running
time. Following this approach, \cite{cai2018efficient} also offsets designing and training each child network from
scratch during the exploration of the highly inefficient architecture search space, by exploring the
architecture space based on the current network and reusing its weights with a bidirectional tree-structured
reinforcement learning meta-controller. This allows for highly expressive tree-structured architecture space which can
be traversed in a multi-branch crawl yielding child architectures in an ordered fashion. Since architecture search
involves efficient distribution of hardware resources and managing close synchronization between the controller and
child networks processes, \aup is effectively used to automate the process. In the remaining section we describe how we
extend and incorporate this algorithm with \aup using the publicly available code~\footnote{From
\url{https://github.com/han-cai/EAS} with commit \code{070d2d7}.}.

The structure of our implementation includes two main parts, \code{client.py} and \code{EASProposer.py}. The
\code{client.py} file is a minor modification of the original file that trains child neural architectures as \ijob{s}.
Illustrated in \autoref{lst:client_old}, the original code takes a folder name as input, which contains the architecture
of the current branch, saved weights, new architecture and a static configuration including new epochs to run and learning
rate. It runs the child net architecture and returns the net validation accuracy and running time. The modified
version (shown in \autoref{lst:client_new}) modifies merely five lines to make it compatible with the \aup framework.
Then \aup handles the execution of these client processes automatically and compiles their results asynchronously to aid
the \iproposer.

\begin{lstlisting}[caption=Origin Client.py,label={lst:client_old},numbers=left,float,xleftmargin=2em]
  from expdir_monitor.expdir_monitor import ExpdirMonitor
  def run(expdir):
    expdir_monitor = ExpdirMonitor(expdir)
    valid_performance = expdir_monitor.run(pure=True, restore=False)
    
  def main():
    expdir = input().strip('\n')
    run(expdir)

  if __name__ == "__main__":
    main()
\end{lstlisting}

\begin{lstlisting}[caption=Updated client.py,label={lst:client_new},float,numbers=left,xleftmargin=2em,
  linebackgroundcolor={
  \ifnum\value{lstnumber}=1\color{yellow}\fi
  \ifnum\value{lstnumber}=3\color{yellow}\fi
  \ifnum\value{lstnumber}=7\color{yellow}\fi
  \ifnum\value{lstnumber}=10\color{yellow}\fi
  \ifnum\value{lstnumber}=11\color{yellow}\fi}
  ]
  #!/usr/bin/env python
  from expdir_monitor.expdir_monitor import ExpdirMonitor
  from aup import BasicConfig, print_result
  def run(expdir):
    expdir_monitor = ExpdirMonitor(expdir)
    valid_performance = expdir_monitor.run(pure=True, restore=False)
    print_result(valid_performance)

  def main():
    config = BasicConfig().load(sys.argv[1])
    run(config["expdir"])
  
  if __name__ == "__main__":
    main()
\end{lstlisting}

The \iproposer wraps the main controller process (\code{arch\_search\_convnet\_net2net.py}), which contains the
RNN-controller based reinforcement critic. Its controlling variables include number of batches to run per episode,
number of episodes to run, maximum epochs per child episode, range of filter dimensions, range of strides, potential
kernel sizes, among many others. The \iproposer is initialized with a basic initial configuration to generate a set of
potential child processes to run. The \aup executes \ijob{s} using the modified \code{client.py} with these
configurations of client networks, and reports back to the original controller once all the generated child nets for the
episode have finished running. The \iproposer then computes gradients from the string of child architectures and the
reported accuracies, and generates new child nets for the next episode using its actors for wider and deeper
configuration generation to build upon the current branch, allowing \aup to take over and execute batches for the
episode iteratively. Once finished, \iuser{s} can easily check the \aup logs and database for the client runs, their
architectures and accuracies, and gain more insights into the experiment. To conclude, the flexibility of \aup design
allows us to easily and quickly integrate an open-source \ac{NAS} code into the framework. 

AutoKeras \cite{jin2019auto} is an open-source library for automated machine learning, and has recently become
increasingly popular for \ac{NAS} applications and research. The library includes a framework and different functions to
search architecture space and hyperparameters for deep learning models. As the early NAS techniques gained popularity,
their major shortcoming of exorbitant computational cost remained unaddressed. In contrast to those techniques,
AutoKeras performs Network Morphism based architecture generation guided by Bayesian Optimization. Network Morphism
keeps the functionality of the neural network while changing its neural architecture, using an edit-distance neural
network kernel which measures how many operations are needed to change one neural network to another. This allows
AutoKeras to minimize the prohibitive computation costs while also allowing for control over the architecture search
space. The framework also provides support for other standard search algorithms like Random, Grid, and Greedy along with
Bayesian Optimization for network morphism. AutoKeras can also be used by NAS researchers who seek to implement their
own NAS algorithms by reimplementing the `generate' and `update' functions to generate the next neural architecture and
update the controller with evaluation result of a neural architecture respectively. 

We provide a high level integration for AutoKeras with \aup, which allows AutoKeras code to be executed on available
resources. \aup takes the `time limit' and `search' as arguments for how long to perform NAS and which search algorithm
to run; and then performs the final Hyperparameter tuning. Our integration allows \iuser{s} to abstract away not only
the NAS search process with AutoKeras but also utilize resource adaptability and result tracking with \aup. Our
AutoKeras integration is designed for both easy scaling on resources for NAS applications  and hassle-free
comparisons for switching between different search techniques or developing new search algorithms. To this end, we
treat each complete AutoKeras search and final tuning as a unique \ijob, unlike our EAS implementation of a granular
approach where each candidate child model would be a \ijob. 

%%%%%%%%%%%%%%%%%%%%%%%%%%%%%%%%%%%%%%%%%%%%%%%%%%%%%%%%%%%%%%%%%%%%%%%%%%%%%%%%%%%%%%%%%%%%%%%%%%%%%%%%%%%%%%%%%%%%%%%%
% Discussion and Conclusion
%%%%%%%%%%%%%%%%%%%%%%%%%%%%%%%%%%%%%%%%%%%%%%%%%%%%%%%%%%%%%%%%%%%%%%%%%%%%%%%%%%%%%%%%%%%%%%%%%%%%%%%%%%%%%%%%%%%%%%%%
\section{Discussion and Conclusion}\label{sec:conclusion}

\aup design goals are focused on a user-friendly interface. \aup benefits both practitioners and researchers and its
design simplifies the integration and development of \ac{HPO} algorithms.  Specifically, the framework design helps both
\iuser{s} to easily use \aup in their workflows and \iresearcher{s} to quickly implement novel \ac{HPO} algorithms.
To reach these goals, the \aup design has fulfilled the following requirements:

\begin{itemize}
  \item Flexibility.  All implemented \ac{HPO} algorithms share the same interface. This enables \iuser{s} to switch
  between different algorithms without changes in the \icode.  A pool of \ac{HPO} algorithms is integrated into the \aup
  for \iuser{s} to explore and for \iresearcher{s} to benchmark against.
  \item Usability. Changes to existing \iuser{'s} \icode are limited to a minimal level. It reduces the friction for
  \iuser{s} to switch to the \aup framework.
  \item Scalability. \aup can deploy to a pool of computing resources to automatically scale out the \iexp, and
  \iuser{s} only need to specify the resource.
  \item Extensibility. New \ac{HPO} algorithms can be easily integrated into the \aup framework if they followed the
  specified interface (see \autoref{sec:proposer}).
  % \item  To support reproducibility, all \ijob records are stored in a persistent database.
\end{itemize}

\aup addresses a critical missing piece in the application aspect of \ac{HPO} research.  It provides a universal
platform to develop new algorithms efficiently.  More importantly, \aup lowers the barriers for data scientists in
adopting \ac{HPO} into their practice.  Its scalability helps users to train their models efficiently with all computing
resources available. Switching between different \ac{HPO} algorithms is simple and only needs changing the proposer name
(dedicated controlling parameters will be default and specified).  This allows practitioners to quickly explore their
ideas with advanced algorithm less laboriously.  

% \aup helps users to tune models with \ac{HPO} into at scale. In this release, it supports both multi-CPU/GPU
% locally and multi-node/AWS resources remotely.  Once \aup is set up, users can simply specify the type and number of
% resources to use, without worrying about job executions anymore.  Furthermore, cloud services or on-premise resource
% managing tools can be easily integrated under the defined abstractions \autoref{sec:rm}. 

% More importantly, \aup lowers the barriers for data scientists to adopt \ac{HPO} into their practice.  Switching between
% different \ac{HPO} algorithms can be simple as only changing the proposer name (dedicated controlling parameters will be
% default and specified).  This allows practitioners to quickly explore their ideas with advanced algorithm less
% laboriously.

The \aup framework requires only minimal changes to existing scripts and these scripts, once modified, can be reused for
other occasions directly.  This non-intrusiveness frees users from repeated refactoring of their code. Users are also
free to use any languages in addition to Python (although a little extra work is needed to setup the interfaces with
\aup).  Altogether, \aup gives practitioners and researchers great flexibility in building models using different
frameworks (e.g. TensorFlow or PyTorch) and multiple languages (e.g. \textsc{MatLab} or \textsc{R}).  We plan to
introduce other functionalities (such as model compression) in \aup in future releases.

To conclude, we have presented the design of \aup that addresses the challenges in current \ac{HPO} solutions. We have
shown that it is user-friendly for both model tuning and new \ac{HPO} algorithms development. \aup supports a few major
\ac{HPO} approaches out of the box\footnote{\textsc{BOHB} and AutoKeras are not included in the version 1
release.} and is ready to help users to automate and accelerate their model training process. We encourage community
contributions to further improve the framework with state-of-the-art algorithms and infrastructure support to solve the
challenges in the big data era.

\bibliographystyle{abbrv}
\bibliography{ref}

\end{document}